# Multi-Site Real-World Performance of Commercial AI for Pulmonary and Incidental Pulmonary Embolism Detection

Aawez Mansuri[1*], Mohammadreza Chavoshi[1*], Theodorus Dapamede[1], Wasif Bala[1], Beatrice Brown-Mulry[1], Rohan Isaac[1], Bardia Khosravi[2], Hanzhou Li[1], Frank Li[1], John T. Moon[1], Chad Robichaux[3], Dan I.G. Cohen-Addad[1], Ninad V. Salastekar[1], Janice Newsome[1], Judy W. Gichoya[1], Hari Trivedi[1#]

[1]Department of Radiology and Imaging Sciences, Emory University School of Medicine, Atlanta, GA, USA
[2]Department of Radiology and Biomedical Imaging, Yale University New Haven, CT, USA
[3]Department of Biomedical Informatics, Emory University, Atlanta, GA, USA

*Aawez Mansuri and Mohammadreza Chavoshi contributed equally and share first authorship.

# Corresponding author: Hari Trivedi (hari.trivedi@emory.edu)

## Abstract

Pulmonary embolism (PE) is a leading cause of cardiovascular mortality, yet the real-world performance of FDA-cleared AI detection models remains incompletely characterized. We retrospectively evaluated two FDA-cleared AI algorithms from a single commercial platform (Aidoc Medical BriefCase) — one for PE triage on dedicated CT pulmonary angiography (CTPA; n = 30,678) and one for incidental PE (iPE) detection on routine contrast-enhanced CTs (n = 37,191), across a 17-facility academic health system. Reference-standard labels were extracted from radiology reports using a validated LLM pipeline (97% accuracy, $\kappa = 0.94$). The PE model achieved 86.8% sensitivity and 99.1% specificity, with sensitivity declining from 99.3% for saddle emboli to 72.9% for subsegmental PE, and from 89.7% for acute to 65.3% for non-acute PE. The iPE model achieved 73.5% sensitivity and 99.8% specificity. Both models demonstrated lower sensitivity than FDA-clearance benchmarks while exceeding cleared specificity, with diminishing performance for peripheral and non-acute emboli mirroring known human reader limitations and underscoring the need for standardized post-market surveillance of AI-enabled medical devices.

## Introduction

Pulmonary embolism (PE) represents a significant public health problem and is the third leading cause of cardiovascular death in the United States[1], affecting as many as 900,000 individuals and resulting in an estimated 60,000 to 100,000 deaths annually[2]. Large pulmonary emboli have an extremely high incidence of mortality with approximately one-quarter of affected individuals experiencing sudden death[3,4]. Importantly, PE does not occur exclusively in symptomatic, high-acuity presentations — incidental pulmonary embolism (iPE), detected on contrast-enhanced CT studies performed for unrelated indications, represents a clinically significant and frequently overlooked entity[5–8]. iPE is often missed because radiologists cease searching for additional pathology once a primary finding is identified, underscoring the need for systematic detection strategies across both dedicated and non-dedicated imaging contexts.

Prompt identification of PE — whether symptomatic or incidental — is critical to preventing adverse outcomes, yet diagnostic delays remain prevalent, affecting nearly 60% of patients[9–11]. Although CTPA is the primary diagnostic modality for suspected PE, the rising volume and required turnaround times for these studies have strained radiological capacity. AI-assisted triage systems have emerged as a promising solution, capable of automatically screening CTPA scans and prioritizing positive findings for immediate radiologist review[12–15]. Recent evidence further suggests that AI can serve as a diagnostic safety net for iPE, improving radiologist sensitivity from 80.0% to 96.2% and reducing report turnaround times[7].

Numerous deep-learning models have been developed to aid in PE detection, with a meta-analysis reporting a pooled sensitivity of 89.4% and specificity of 87.1%[16]. While several models have achieved FDA clearance[17–19] and demonstrated promising retrospective performance, their real-world clinical utility remains incompletely characterized. Existing evaluations on real-world data report a wide range of performance — sensitivities from 63.0% to 96.8% and specificities from 95.0% to 99.9%[15,20] — and critical gaps persist in evaluation across diverse patient settings (inpatient, outpatient, emergency), anatomical thrombus locations and subtypes, and demographic subgroups. Such evaluations are essential, as the clinical value of AI depends on its ability to augment radiologist performance in subtle or easily overlooked cases, not merely those that are readily apparent.

In this study, we assess the performance of a commercial PE triaging model (Aidoc Medical, NY), an FDA-cleared AI system designed for automated PE detection on CTPA imaging. Per its FDA 510(k) clearance, the model achieved a sensitivity of 94.86% (95% CI: 90.99%–97.41%) and specificity of 94.04% (95% CI: 90.62%–96.49%) on a validation set of 499 cases[21]. However, the 510(k) summary provides no information on embolus location, acuity, laterality, patient demographics, or clinical setting — gaps that our study directly addresses.

Using two years of data from a large academic medical center, we provide a comprehensive retrospective analysis of model performance across demographic, clinical, and anatomical subgroups on dedicated CTPA exams. Both algorithms evaluated here are components of a single commercial platform (Aidoc Medical BriefCase) rather than two competing products, and this study is therefore not a head-to-head comparison of different vendors. We simultaneously conduct a parallel evaluation of a second FDA-cleared AI model for automatic iPE detection on routine, non-angiographic Chest and Abdominal CTs[22]. This dual approach characterizes these models' utility across the full spectrum of real-world PE detection.

## Results

### Patient Cohort

A total of 40,490 dedicated CTPA examinations and 50,362 routine contrast-enhanced chest and abdomen CT studies were initially identified across the study period. From the primary CTPA cohort, 1,685 examinations with indeterminate PE classification and 8,127 reported as suboptimal were excluded, yielding a final primary cohort of 30,678 unique studies from 26,288 patients. These exclusion labels were derived from the language of the final radiology report using the same validated LLM pipeline: an examination was classified as indeterminate when the report used hedging language regarding the presence of PE (e.g., a PE that could be neither confidently confirmed nor excluded), and as suboptimal when the interpreting radiologist explicitly qualified that the examination was of inadequate quality for confident PE assessment. These designations follow standard radiologic reporting conventions at our institution and were not created or modified for this study. Suboptimal quality most commonly reflected limited contrast bolus timing, followed by image artifacts and respiratory motion; it does not refer to the number of acquired slices, as all imaging protocols use standardized thin-section reconstructions. A sensitivity analysis of cases excluded on the basis of ambiguous ground truth is provided in Table S10. From the routine CT cohort, 7,594 examinations were excluded based on temporal proximity to dedicated PE studies (within 7 days prior to or 24 hours after the routine CT examination), to ensure the iPE findings were truly incidental rather than follow-up assessments of known or suspected PE. An additional 5,496 examinations were excluded when patients underwent multiple procedure types within a 24-hour period, with preference given to retaining chest CT examinations over abdomen-pelvis studies to optimize pulmonary vascular visualization. This de-duplication step applied only to the secondary (iPE) cohort and refers to CT examinations: when a patient underwent more than one qualifying contrast-enhanced CT within a 24-hour window, a single examination was retained per patient to avoid counting the same incidental finding more than once, with chest CT preferred over abdomen-pelvis CT because the pulmonary vasculature is more completely and reliably visualized on chest acquisitions. Furthermore, 81 cases that were labeled as 'Possible' by LLM were also dropped from the evaluation. These 81 'Possible' iPE cases correspond to reports in which the presence of incidental PE could not be confidently established, and were excluded from the primary evaluation because the AI model produces only a binary output and no verifiable ground truth could be assigned. We note that this 'Possible' designation in the secondary cohort is distinct from the 'Present/Possible' associated-finding labels reported for pulmonary infarction, right heart strain, and pulmonary arterial hypertension in the primary cohort (Table 4), and from the iPE label-extraction validation summarized in Figure S5. This yielded a final iPE cohort of 37,191 unique studies from 28,226 patients. Demographic characteristics of both cohorts are summarized in Table 1, and the full cohort selection process is illustrated in Figure 2.

**LLM Performance on PE Label Extraction**

LLM-based label extraction demonstrated strong overall performance across both cohorts, achieving an accuracy of 97% and a Cohen's kappa of 0.94 against the expert reference standard. The confusion matrix revealed excellent agreement across positive and negative classes, with the majority of discrepancies confined to the indeterminate category (Figure S1). Excluding indeterminate ("Possible") cases, the LLM achieved a sensitivity of 100%, specificity of 92%, accuracy of 97%, PPV of 96%, and NPV of 100% for PE detection on dedicated CTPA reports (Table S4), with comparable binary classification performance applied to routine CT reports for iPE identification (Figure S5).

Performance remained high across all sub-labels on the primary cohort (Table S4). Extraction of anatomical embolism depth yielded sensitivities of 91–100% and specificities of 76–100%. For acuity classification, the LLM achieved 100% sensitivity and 97% specificity for acute PE, and 86% sensitivity with 97% specificity for non-acute PE. Laterality extraction was similarly robust, with accuracies of 99–100% and Cohen's kappa of 0.97–1.00 across right, left, and bilateral classifications. For associated clinical findings, the LLM achieved sensitivity/specificity of 97%/99% for right heart strain, 93%/99% for pulmonary hypertension, and 100%/98% for pulmonary infarction.

**Distribution of Labels in PE cases**

Of the 30,678 dedicated CTPA examinations in the primary cohort, 3,279 (10.7%) were positive for PE. The majority were acute (88.1%), with non-acute PE comprising the remaining 11.9%. Bilateral involvement was most common (47.3%), followed by right-sided (37.9%) and left-sided (13.6%) disease. At the most proximal level of thrombus involvement, segmental PE was most frequent (40.1%), followed by main/lobar (26.6%) and saddle emboli (4.5%). Associated findings included pulmonary infarction (23.5%) and pulmonary hypertension (18.8%). In the secondary cohort of 37,191 routine contrast-enhanced studies, 538 cases (1.4%) were identified as incidental PE on the original radiology report, a count that includes the 81 cases labeled 'Possible'; after exclusion of these indeterminate cases, 457 confirmed iPE cases remained available for diagnostic performance evaluation. Returning to the primary cohort, The prevalence of PE-positive findings varied significantly across patient class, with differences between emergency and outpatient settings (e.g., 89.3% vs 82.1% acute PE), with additional differences observed by race, age, and sex across both cohorts (Table 2, Table S5).

**AI Model Performance**

For dedicated PE detection on CTPA, the model achieved a sensitivity of 86.8%, specificity of 99.1%, accuracy of 97.8%, PPV of 92.3%, NPV of 98.4%, and F1-score of 0.895. For incidental PE detection on routine contrast-enhanced studies, the dedicated iPE algorithm achieved a sensitivity of 73.5% and specificity of 99.8%, correctly flagging 336 of 457 confirmed iPE cases (the 457 confirmed cases represent the 538 iPE-positive reports after exclusion of the 81 'Possible' cases). The demographic composition of the iPE cohort is summarized in Table S11, and iPE model performance across demographic subgroups (patient class, race, sex, age, and ethnicity) is reported in Table S12. Anatomical subtype-level performance (e.g., embolus location, laterality, right heart strain) was evaluated for the PE model but not for iPE, because these features cannot be reliably ascertained on non-dedicated examinations and are therefore not consistently described in iPE reports; the overall low prevalence of iPE further limited finer stratification. The model performance was also evaluated on 500 radiologists-annotated exams (Table S3).

**Demographic Subgroup Performance for PE**

Full demographic subgroup results are provided in Table 3. For dedicated PE detection, model sensitivity was lowest in outpatients (83.7%, 95% CI: 78.5–88.6%) compared with inpatients (86.5%, 95% CI: 84.3–88.6%) and emergency department patients (87.3%, 95% CI: 85.8–88.8%). Sensitivity was slightly higher among White patients (88.5%, 95% CI: 86.5–90.4%) compared to Black (85.9%, 95% CI: 84.4–87.5%) and Hispanic patients (83.6%, 95% CI: 76.3–89.3%). By sex, sensitivity was 87.1% (95% CI: 85.5–88.7%) in females and 86.4% (95% CI: 84.6–88.3%) in males. Across age groups, sensitivity was highest in patients aged 65 and older (88.0%, 95% CI: 86.4–89.7%), followed by those under 40 (87.4%, 95% CI: 84.2–90.6%) and patients aged 40–65 (85.1%, 95% CI: 83.1–87.1%). To assess whether the higher sensitivity in the 65+ group was driven by the oldest patients, we examined finer age strata (Table S14); sensitivity was stable within the 65+ group (87.5% for 65–75 years and 88.5% for

≥75 years) and across all strata (range 84.7–88.5%), indicating that the age-group findings were not attributable to differential performance in the oldest subgroup.

### PE Subtype Performance

Model performance for PE detection varied substantially by embolism characteristics (Table 4). Sensitivity was high for acute PE (89.7%, 95% CI: 88.6–90.8%) but dropped markedly for non-acute cases (65.3%, 95% CI: 60.7–69.9%). A strong gradient was observed across anatomical depth: sensitivity was near-perfect for saddle emboli (99.3%) and progressively declined for main/lobar (91.2%), segmental (82.3%), and subsegmental (72.9%) PE. Bilateral PE was detected with the highest sensitivity (95.2%), while left-sided unilateral PE showed the lowest (70.5%). The presence of secondary signs of clot burden further influenced performance; sensitivity was 95.2% in cases with right heart strain versus 82.2% without, and 90.9% in cases with pulmonary infarction versus 85.5% without. Representative false-positive, false-negative, and true-positive cases, including Grad-CAM visualizations of model attention, are shown in Figures S2, S3, and S4, respectively.

### Intersectional Demographic and Clinical Setting Analysis

Intersectional analysis revealed meaningful performance variation across combinations of age and patient setting (Table S6). The most pronounced gap was observed in middle-aged patients (40–65), where outpatient sensitivity (77.1%, 95% CI: 67.4–86.2%) was markedly lower than in emergency (86.3%) and inpatient (84.2%) settings. Among older adults (65+), sensitivity was stable across all settings (87.5–88.3%). In patients under 40, outpatient sensitivity appeared highest (94.4%); however, this estimate is based on only 18 cases (95% CI: 81.8–100%) and should be interpreted with caution. Specificity remained consistently high across all intersectional subgroups, generally exceeding 98.7%. Corresponding intersectional analyses by age and race, and by patient setting and race, are provided in Tables S7 and S8, respectively.

### Multivariate Analysis

Multivariable logistic regression identified embolism acuity, anatomical level, and the presence of right heart strain as features associated with correct model prediction after mutual adjustment (Table S9). Non-acute PE was associated with a 79% reduction in the odds of correct detection compared to acute PE (OR 0.21, 95% CI: 0.16–0.27). Relative to segmental emboli, the odds of correct detection were substantially higher for saddle (OR 15.33, 95% CI: 2.06–114.03), main (OR 3.59), and lobar (OR 2.03) emboli, and significantly lower for subsegmental PE (OR 0.54, 95% CI: 0.40–0.73). The presence of right heart strain was associated with nearly three-fold higher odds of correct detection (OR 2.73, 95% CI: 1.99–3.74). Pulmonary arterial hypertension and pulmonary infarction were not independently associated with model performance. The very wide confidence interval for saddle emboli reflects the small number of saddle cases and the near-uniformly correct detection within this group; the corresponding odds ratio should therefore be interpreted as evidence of direction rather than a precise effect size.

**Clinical Impact Analysis**

To assess the clinical relevance of iPE detections, we analyzed 1,225 patients in the secondary cohort who underwent a dedicated PE-protocol examination within 30 days of an index routine contrast-enhanced CT, comparing the AI flag generated at the index scan against PE outcomes on the dedicated follow-up study (Figure 3). On the index radiology report, 91 of the 1,225 examinations (7.4%) were iPE-positive or possible, while 1,134 (92.6%) were iPE-negative. Among the 91 index iPE-positive/possible cases, the algorithm flagged 54 (59.3%) and missed 37 (40.7%). Sixty-one of these 91 cases were confirmed to have PE on dedicated follow-up imaging; of these 61 confirmed PEs, the algorithm had already flagged 42 (42/61, 68.9%) on the index non-dedicated scan and missed 19 (19/61, 31.1%). Viewed by the algorithm's own output, 42 of the 54 flagged cases (42/54, 77.8%) and 19 of the 37 missed cases (19/37, 51.4%) were confirmed to have PE on follow-up. Among the 1,134 index iPE-negative examinations, the algorithm produced 7 false-positive flags (0.6%), with the remaining 1,127 (99.4%) concordant with the negative index report. Taken together, the algorithm flagged 42 of the 61 incidental PEs ultimately confirmed on dedicated follow-up imaging (68.9%) at the index scan, while roughly one-third (19/61, 31.1%) were not flagged. In this clinically selected subset the algorithm therefore captured most, but not all, confirmed incidental PEs, indicating that it functions as an adjunct to—rather than a replacement for—radiologist interpretation, and that continued radiologist vigilance remains essential.

**Discussion**

Our results reveal distinct performance profiles for both the PE and iPE models relative to manufacturer-reported FDA 510(k) metrics. For PE detection, our observed sensitivity of 86.8% was lower than the 94.9% reported in the FDA submission, while our specificity of 99.1% substantially exceeded the cleared 90.1%[21]. This pattern is consistent with published literature, where reported sensitivities for this model have ranged from 63.0% to 96.8% while specificity has remained consistently high[14,20,24–26]. A 2025 meta-analysis of FDA-approved AI algorithms for PE detection corroborated this, reporting pooled sensitivity and specificity of 0.91/0.94 during internal validation, declining to 0.89/0.88 externally—confirming that performance degradation upon real-world deployment is a reproducible phenomenon [27]. For the iPE model, our sensitivity of 73.5% fell below the FDA-reported 89.7%, while specificity of 99.8% exceeded the cleared 90.1%[22]—mirroring the PE model's pattern and falling between the 50.0% and 96.4% reported in recent external evaluations[7,28]. These discrepancies are likely multifactorial, reflecting differences in model version, patient population, comorbidity burden, and imaging protocol heterogeneity typical of large academic centers. We were unable to independently verify that the model version deployed during our study period was identical to the specific version evaluated in the manufacturer's 510(k) submission; the cleared algorithm may have been updated between clearance and deployment, and any such version difference could contribute to the observed performance gap. In addition, the manufacturer's reported metrics were derived from a curated validation cohort with its own intended-use population, operating point, and inclusion criteria, which differ from our real-world cohort in case mix, disease prevalence, and protocol heterogeneity. The comparison with FDA-clearance metrics should therefore be regarded as illustrative of the gap between controlled validation and real-world deployment rather than as a like-for-like benchmark.

These performance gaps reflect a deeper structural issue in the regulatory pathway for AI/ML medical devices. A 2025 systematic review found that 97% of FDA-cleared radiology AI devices were approved via 510(k), with only 5% undergoing prospective testing, 29% incorporating any clinical testing, and fewer than 2% supported by randomized controlled trials—with nearly half of FDA decision summaries omitting study design or sample size entirely[29]. Our findings directly illustrate this gap: models cleared under controlled conditions demonstrated meaningfully different sensitivity in a large, heterogeneous real-world cohort. In our view, reporting aggregate binary classification metrics alone is insufficient for characterizing such devices, because pooled sensitivity and specificity can mask substantial heterogeneity across clinically meaningful subgroups—a model that performs well chiefly on findings human readers already identify easily may offer limited additive value where it is most needed. This argues for subgroup-level performance reporting as a routine component of validation rather than for any change in which summary metrics are computed.[36–38] Our results therefore support the growing call for mandatory, standardized post-market surveillance of AI-enabled medical devices, a position now endorsed by both the European Society of Radiology and emerging U.S. regulatory guidance[30,31].

Both models exhibit performance gradients tied to finding severity and anatomical location. The PE model achieves highest sensitivity for acute, central emboli and cases with right heart strain, with sensitivity declining progressively for peripheral and subsegmental PE. Sensitivity was also lower for isolated left-sided PE, potentially reflecting underrepresentation of left-sided cases in training data, consistent with their lower prevalence in the general population[32]. This gradient mirrors known human radiologist performance limitations. Rather than viewing this as a simple failure, the literature increasingly frames AI utility as greatest where human performance is most variable—a 2024 study found that AI assistance benefited lower-performing radiologists more than high performers, suggesting that targeting AI development toward harder findings maximizes clinical impact[33]. Beyond illustrative examples, these subgroup and multivariable analyses constitute a systematic, cohort-wide failure analysis of the model: rather than characterizing errors through a handful of hand-picked cases, they quantify where detection degrades across the full deployment population and identify the imaging features most strongly associated with missed detection. A deeper, feature-level failure analysis—an automated, LLM-driven post-deployment monitoring pipeline that extracts a broad set of concurrent imaging findings from reports and models their association with model errors—is a natural extension of this work and is planned as a dedicated downstream study.

The iPE algorithm's clinical value must also be understood in the context of the diagnostic environment in which it operates. Unlike dedicated PE-protocol CTPA, non-PE protocol examinations direct clinical attention elsewhere, making radiologists susceptible to "satisfaction of search" and inattentional blindness—with reported miss rates as high as 45.5% in specialized studies such as cardiac CTs[34].In this context, even a sensitivity of 73.5% represents meaningful additive value, as the algorithm functions as a systematic second-pass review across every eligible examination regardless of the radiologist's primary focus. However, we found that the model missed 31.1% of iPEs detected by radiologist, highlighting the need for continued vigilance by radiologists.

Performance was broadly consistent across age, sex, and patient class. Sensitivity was slightly lower in some minority subgroups (e.g., Asian patients: 85.21%, 95% CI 74.60%–93.75%), though wide confidence intervals preclude definitive conclusions. Regulatory bodies and professional societies have increasingly called for disaggregated subgroup reporting as a standard component of both pre-market submissions and post-market surveillance, recognizing that training data imbalances can produce systematic real-world disparities[30,35]. Our subgroup-level reporting with confidence intervals directly responds to this expectation and offers a methodological template for future evaluations.

This study is not without limitations. As a single large academic center study, generalizability is limited; performance may differ in community or rural settings with different patient populations, imaging protocols, and hardware, necessitating multi-site prospective validation of both algorithms. Our reference standard was derived from final radiology reports rather than independent blinded image adjudication. Because the AI model was clinically active during the study period, the report-derived labels are not independent of the AI output, and our results should therefore be interpreted as model performance against AI-exposed clinical reports rather than against an independent expert image-based reference standard. The most likely direction of this bias favors the model: a radiologist with access to the AI flag is more likely to confirm a true but subtle embolus that might otherwise have been overlooked, which would tend to increase the apparent true-positive rate, whereas it is implausible that a radiologist who identifies a PE would be dissuaded from reporting it by a negative AI result, since for PE the finding is either present or absent on the image rather than inferred from ambiguous features. Conversely, cases the radiologist still reported as negative despite a positive AI flag remain counted as false positives, so the specificity estimate is not inflated by this exposure. Blinded image re-adjudication of discordant cases was not feasible within the scope of this study given its scale, and, more fundamentally, single-reader re-review would not yield a definitive reference standard for the subtle and hedged cases at issue, as inter-radiologist agreement on such cases is itself limited and would require a dedicated multi-reader consensus study. LLM-based label extraction, while validated and scalable, introduces potential mislabeling risk for ambiguous or hedged report language, though cross-validation against a manually labeled 500-case subset demonstrated consistent subgroup trends. Subgroup analyses across racial and ethnic minority groups were constrained by sample size, with wide confidence intervals precluding definitive conclusions about differential performance. Finally, as a retrospective study, we cannot establish causal relationships between AI flagging and clinical outcomes such as time-to-treatment or mortality.

In conclusion, our real-world evaluation demonstrates that both a commercial PE and iPE models exhibit significantly variable performance in real-world deployment compared to FDA-clearance metrics with decreasing performance for smaller emboli, mirroring known human reader limitations. Subgroup performance was broadly equitable, though definitive conclusions in smaller demographic groups require larger samples. Taken together, these findings underscore that real-world post-deployment evaluation is not an optional institutional exercise but a clinical and ethical obligation—and that the field urgently requires standardized frameworks for continuous AI performance monitoring across diverse patient populations.

## Methods

This retrospective, single-center study was conducted across a 17-facility academic healthcare system and received approval from the institutional review board (IRB: STUDY00002276).

### Patient Cohort

The study population included all patients aged 18 years and older who underwent imaging evaluated by the AI model between April 2023 and April 2025. The AI models are deployed within the routine clinical workflow and run automatically on every eligible examination; there is no patient- or study-level opt-out mechanism, and all eligible studies during the study period were processed by the model. The primary cohort consisted of all Computed Tomography Pulmonary Angiography (CTPA) examinations (n = 40490) performed for suspected PE. A secondary cohort was identified comprising all routine contrast-enhanced chest and abdominal CT examinations (n = 50,362) performed during the same period for indications other than suspected PE, used to evaluate the model's performance in detecting incidental pulmonary embolism (iPE). To ensure findings in the secondary cohort were truly incidental, we excluded any case in which a dedicated PE-protocol study was performed ≤7 days before or ≤24 hours after the index CT, mitigating the risk of radiologist bias from prior knowledge of a PE diagnosis.

**AI Triage Models**

Two FDA-cleared commercial deep learning algorithms (BriefCase, Aidoc Medical, New York, NY) were evaluated in parallel, one for pulmonary embolism (PE) triage and the other for incidental PE (IPE) detection. For the primary cohort, a PE triage model optimized for dedicated CTPA protocols provided a binary, study-level output indicating suspected presence or absence of PE. For the secondary cohort, a separate algorithm designed specifically for incidental PE detection analyzed routine contrast-enhanced chest and abdomen CT examinations. Both systems operate within the same clinical infrastructure: images from the Picture Archiving and Communication System (PACS) are securely routed to a private cloud for AI analysis, with results returned to the radiologist's workstation via a custom user interface. Positive flags generate a notification to the interpreting radiologist. At the workstation, the radiologist views a single, case-level binary flag for each examination. For this study, the vendor provided these same per-exam binary predictions as a combined dataset covering all processed examinations. The exported prediction for each case is identical to the binary flag displayed to the radiologist at the time of interpretation; the combined export was obtained solely to enable efficient retrospective, system-wide analysis rather than because it contained any additional or differently formatted information. Notably, AI model outputs were available to the interpreting radiologist at the time of clinical interpretation for all cases in both cohorts [Figure 2].

**Ground Truth Determination via LLM Extraction**

A unified ground truth labeling workflow was applied across both cohorts using a large language model (LLM)-based extraction pipeline.

All examinations were documented using standardized institutional templates by board-certified attending radiologists during routine clinical care. A comprehensive annotation schema was developed by a consensus panel of three board-certified emergency radiologists. For the primary CTPA cohort, the schema defined clinically relevant labels including study quality (limited/suboptimal), PE presence (positive, negative, or indeterminate), acuity, location (saddle, main, lobar, segmental, subsegmental), laterality (right, left, bilateral), and associated findings such as pulmonary hypertension, right heart strain, and pulmonary infarction (Table S1) [Figure 1]. For the secondary iPE cohort, reports were classified as positive, negative, or indeterminate for PE.

To train and validate the LLM extraction process, a reference dataset of 500 CTPA reports was randomly sampled from the primary CTPA cohort, drawn across all 17 facilities and spanning the same study period. Of these, 300 reports were used as a training set for iterative prompt optimization. The remaining 200 reports were reserved as a held-out validation set and comprised two independently annotated 100-report subsets: one annotated by radiology residents and verified by an attending radiologist, and one adjudicated by three board-certified attending radiologists (Table S2). Because the prompts were finalized on the training set and frozen prior to evaluation, with no modifications made on the basis of either evaluation subset, and because both subsets were annotated under the same schema, the two subsets were combined into a single 200-report validation set for final LLM performance assessment.

The GPT-4o model (OpenAI, San Francisco, CA) was used for label extraction[23]. A zero-shot learning approach was employed with iterative prompt engineering, in which candidate prompts were tested and finalized based on top performance across all labels on the training set; the prompts were then frozen prior to evaluation on the validation set. The optimized prompts were then applied to all radiology reports across both cohorts. For the primary cohort, labels were extracted hierarchically: reports were first classified as PE-positive, negative, or indeterminate, after which PE-positive reports were further annotated for detailed sub-labels (e.g., location, acuity, right heart strain). For the secondary cohort, the same pipeline was applied to extract a binary PE-presence label from routine CT chest or abdomen reports. No separate prompt-engineering was performed for the iPE cohort; the binary PE-presence prompt finalized on the primary CTPA training set was reused verbatim for label extraction from routine chest and abdomen CT reports. Consequently, the entire set of independently annotated iPE reports could be reserved for validation rather than prompt development, and all 250 annotated iPE reports were used to evaluate iPE label-extraction performance (Figure S5).

## Statistical Analysis

LLM extraction performance was assessed by comparing output against the manually annotated 200-report validation set. For binary labels, sensitivity, specificity, and accuracy were calculated. For multi-class labels, a confusion matrix was generated alongside accuracy, F1-score, and Cohen's kappa to assess inter-rater agreement. Label-specific performance with 95% confidence intervals and per-label sample sizes is reported in Table S4, the overall extraction confusion matrix is shown in Figure S1, and the prevalence of each clinical label within the validation set is summarized in Table S13.

The diagnostic performance of both AI models was evaluated against their respective LLM-derived ground truth labels. Exams reported as indeterminate or suboptimal for evaluation were excluded due to ambiguity of ground truth. For each cohort, we calculated sensitivity, specificity, positive predictive value (PPV), negative predictive value (NPV), F1-score, and accuracy. These metrics were further stratified by demographic subgroups for both cohorts. Age strata (18–40, 40–65, and ≥65 years) were pre-specified according to established clinical convention distinguishing young, middle-aged, and older adults, with 65 years as the conventional older-adult threshold, and were not derived from the data; the unequal number of examinations across strata reflects the age distribution of the underlying CTPA population, in which PE prevalence increases with age. For imaging-based subgroups in the primary cohort, sensitivity was calculated independently, as the model does not predict specific PE subtypes. To investigate the influence of correlated imaging features on model predictions in true positive cases, a multivariate logistic regression was performed with the model's binary prediction (correct vs. incorrect) as the dependent variable and imaging features as covariates. The analytic population for this model comprised all PE-positive examinations in the primary cohort, with the outcome defined as whether the AI model correctly flagged each case; examinations with unspecified values for a given feature were not assigned that feature level. The candidate features (embolism acuity, anatomical level, laterality, right heart strain, pulmonary arterial hypertension, and pulmonary infarction) describe clinically distinct attributes of an embolus and were therefore treated as non-collinear covariates rather than as formally orthogonalized predictors. Statistical significance was set at $p < 0.05$. All performance metrics were reported with 95% confidence intervals estimated via non-parametric bootstrap resampling with 1,000 iterations. Resampling was performed at the examination level. Because dedicated CTPA studies separated in time represent independent diagnostic events in which the presence or absence of PE at one timepoint does not determine the result at another, repeated examinations from the same patient were treated as independent observations rather than as correlated repeated measures of a single underlying state.

To assess the clinical utility of AI-flagged iPE, a longitudinal analysis was performed on patients in the secondary cohort who underwent a subsequent dedicated PE-protocol exam within 30 days of the index routine CT. AI model outputs from the index scan were compared against report-derived labels from follow-up dedicated imaging to determine diagnostic concordance within this clinically selected subset. Because patients who undergo dedicated follow-up imaging are selected on the basis of clinical suspicion, symptoms, and report findings, this analysis reflects concordance with follow-up imaging in a selected subgroup rather than a validated estimate of the algorithm's standalone effect on missed-diagnosis rates or patient outcomes.

The statistical analysis was performed in two stages: evaluation of the LLM extraction process and evaluation of the AI triage model's diagnostic performance.

**Data Availability:** The datasets generated and/or analyzed during the current study are not publicly available due to patient privacy protections (protected health information) and institutional data-sharing policy, but are available from the corresponding author on reasonable request.

**Code Availability:** The code generated and/or analyzed during the current study is not publicly available due to institutional policy but is available from the corresponding author on reasonable request.

**Acknowledgements:** This study was funded, in part, the National Institutes of Health (NIH) Agreement No. 1OT2OD032581. The views and conclusions contained in this document are those of the authors and should not be interpreted as representing the official policies, either expressed or implied, of the NIH. We thank Aidoc Medical for providing the aggregated prediction outputs used in this analysis. Aidoc Medical had no role in the study design, data analysis, interpretation of results, or preparation of the manuscript.

Judy Gichoya receives funding from the National Heart, Lung and Blood Institute (NHLBI) grant R01HL167811; R01HL177003) and National Institutes of Health (grant 1OT20D038065-01; 1R25OD039834-01)

**Author Contribution Statement:**

A.M. and M.C. contributed equally to this work and share first authorship. A.M.: Conceptualisation, study design, methodology, data collection, software, validation, formal analysis, writing (original draft), review & editing; accessed and verified the data. M.C.: Conceptualisation, study design, methodology, data collection, data curation, formal analysis, visualisation, writing (original draft), review & editing; accessed and verified the data. T.D.: Data collection, validation, writing (review & editing). W.B.: Data curation, data collection, validation, writing (review & editing). B.B-M.: Investigation, data collection, writing (review & editing). R.I.: Data collection, data curation, writing (review & editing). B.K.: Data analysis, visualisation, writing (review & editing). H.L.: Data curation, formal analysis, writing (review & editing). F.L.: Data collection, validation, writing (review & editing). J.T.M.: Data analysis, methodology, writing (review & editing). C.R.: Data collection, resources, writing (review & editing). D.I.G.C-A.: Data curation, methodology, writing (review & editing). N.V.S.: Formal analysis, validation, writing (review & editing). J.N.: Supervision, methodology, data interpretation, writing (review & editing). J.W.G.: Supervision, methodology, study design, data interpretation, resources, writing (review & editing). H.T.: Supervision, conceptualisation, study design, funding acquisition, project administration, data interpretation, writing (review & editing); accessed and verified the data. All authors reviewed the manuscript.

**Competing Interests:** The authors declare no competing financial or non-financial interests.

Tables and Figures

**Figure 1.** Label extraction hierarchy. Acuity, Depth, Laterality, Pulmonary Artery Hypertension, Right Heart Strain and Pulmonary Infarction were extracted for all PE positive reports.

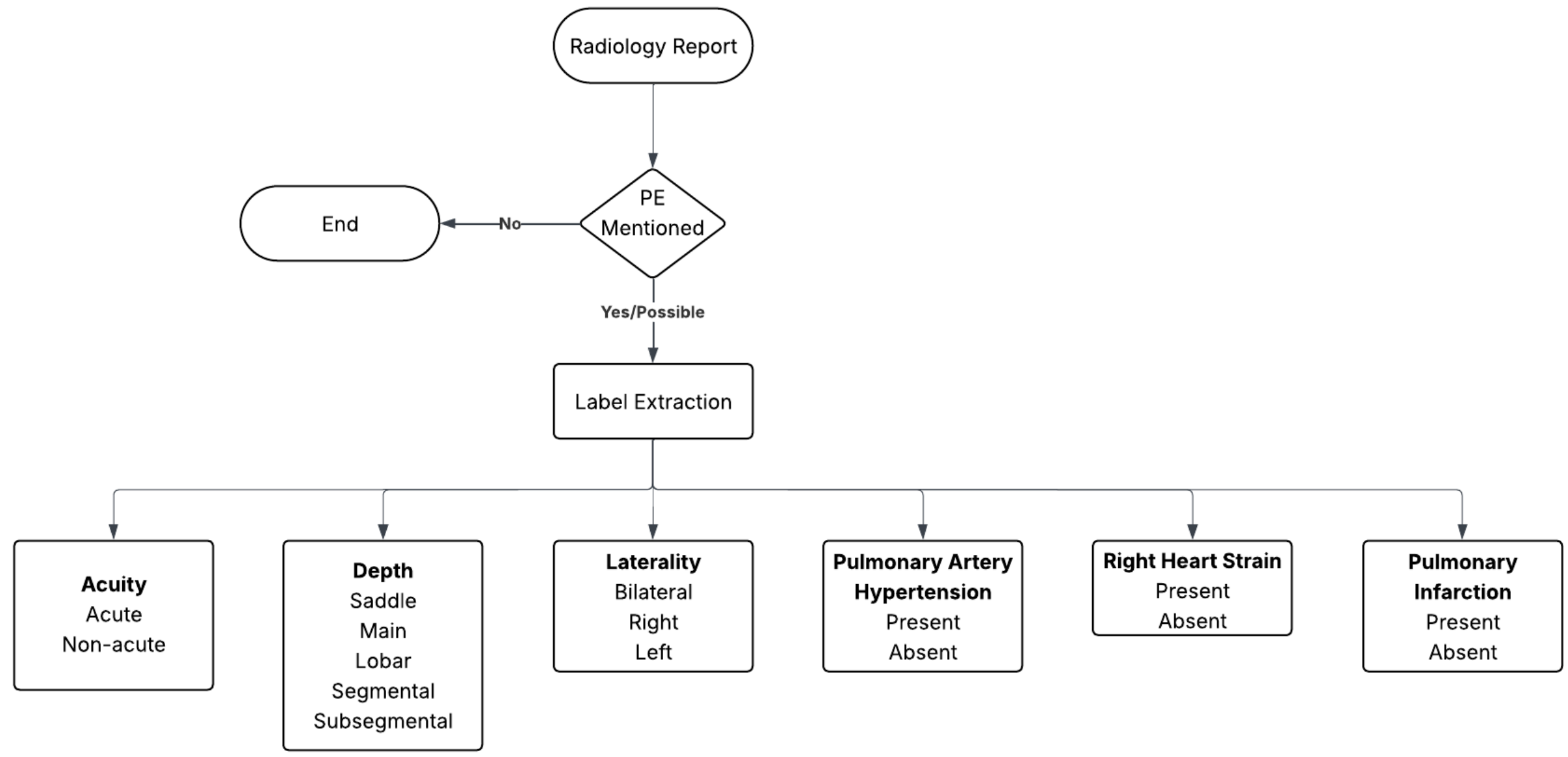

**Figure 2. Workflow of AI model inference and comparison with LLM-extracted labels for our evaluation.**

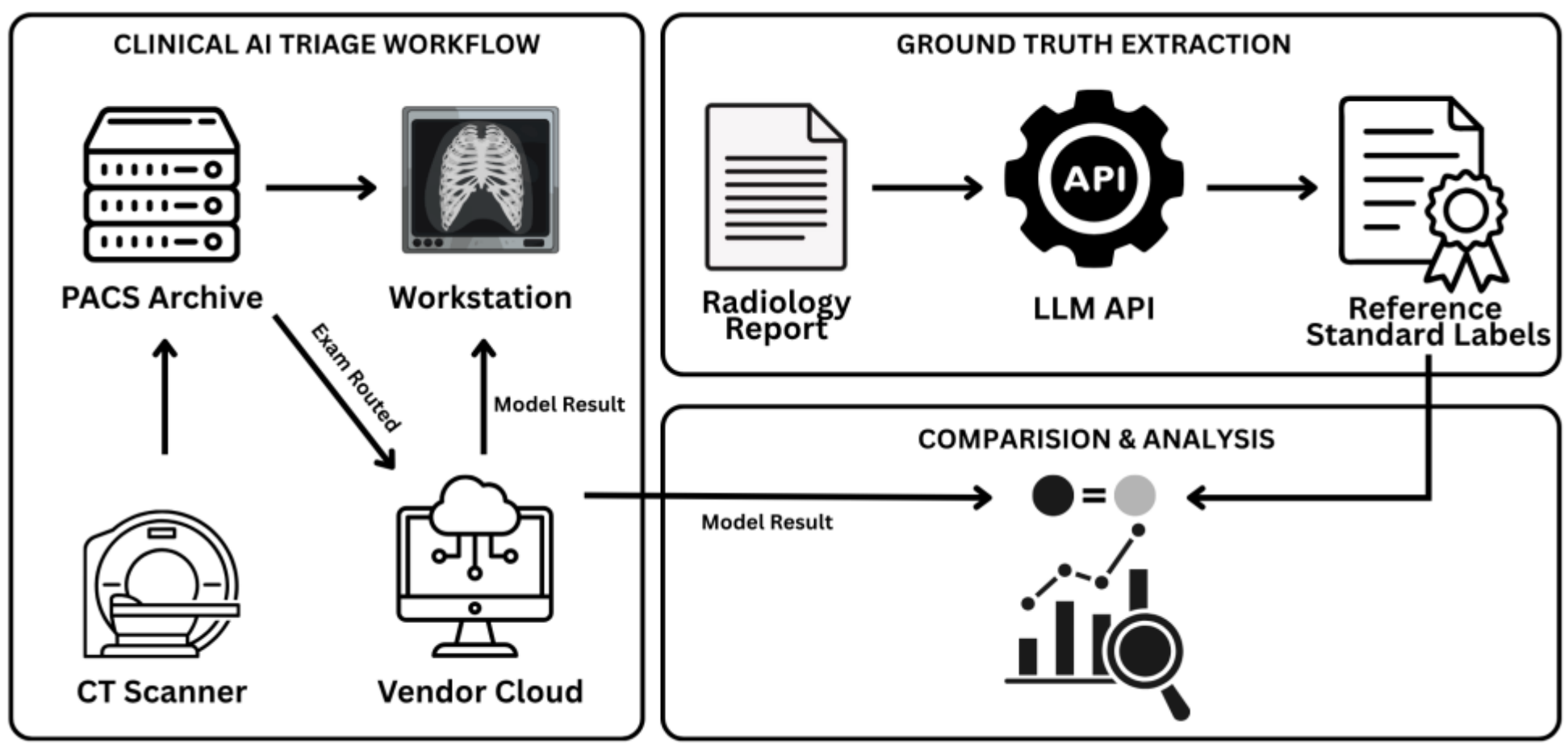

**Figure 3. Clinical validation of the incidental pulmonary embolism (iPE) triage model against dedicated follow-up imaging.**

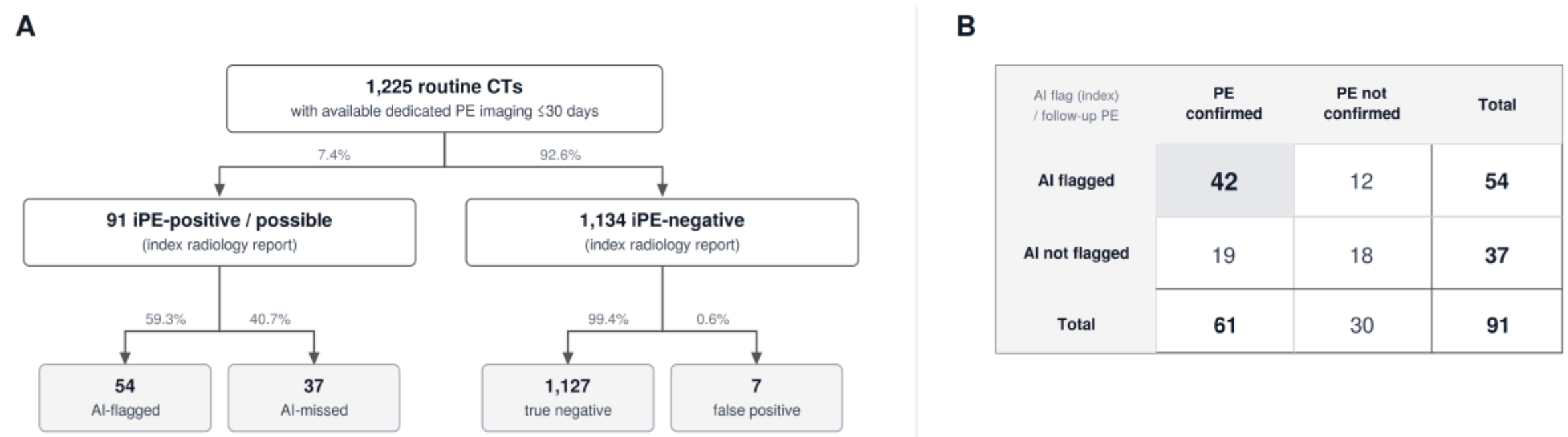

| AI flag (index) / follow-up PE | PE confirmed | PE not confirmed | Total |
|---|---|---|---|
| AI flagged | 42 | 12 | 54 |
| AI not flagged | 19 | 18 | 37 |
| Total | 61 | 30 | 91 |

**(A)** Selection and index classification of the follow-up cohort. Of 1,225 routine contrast-enhanced CTs with an available dedicated PE-protocol study within 30 days, 91 (7.4%) were read as iPE-positive/possible on the index radiology report and 1,134 (92.6%) as iPE-negative. Among the 91 index-positive cases, the algorithm flagged 54 (59.3%) and did not flag 37 (40.7%); among the 1,134 index-negative cases, 1,127 (99.4%) were true negatives and 7 (0.6%) were false positives.

**(B)** Cross-tabulation of the algorithm's index-scan flag against PE outcome on dedicated follow-up imaging, restricted to the 91 index iPE-positive/possible cases. Of these, 61 were confirmed to have PE on follow-up. Reading down the confirmed-PE column, 42/61 (68.9%) of confirmed PEs had been flagged by the algorithm at the index scan and 19/61 (31.1%) had not. Reading across the algorithm's own predictions, 42/54 (77.8%) of AI-flagged cases and 19/37 (51.4%) of AI-missed cases were confirmed to have PE on follow-up. The shaded cell (42) is the shared numerator of the 68.9% and 77.8% rates, which differ only in their denominator (61 versus 54).

**Table 1.** Demographic distribution of included cohort

| Category | Subcategory | Count | Percentage |
|---|---|---|---|
| **Sex** | Female | 18,781 | 61.2% |
| | Male | 11,894 | 38.8% |
| | Not available | 3 | 0.00% |
| **Age** | Mean (SD) | 58.1 (17.9) | – |
| | Median | 60 | – |
| | 18-40 | 5,581 | 18.2% |
| | 40-65 | 12,619 | 41.1% |
| | 65+ | 12,478 | 40.7% |
| **Race** | Asian | 1,123 | 3.7% |
| | Black | 17,747 | 57.9% |
| | Other | 964 | 3.1% |
| | Not available | 857 | 2.8% |
| | White | 9,987 | 32.6% |
| **Ethnicity** | Hispanic | 1,407 | 4.6% |
| | Not Hispanic | 28,018 | 91.3% |
| | Not available | 1,253 | 4.1% |
| **Clinical Service Location** | Emergency Department | 22,084 | 72.0% |
| | Inpatient | 6,403 | 20.9% |
| | Outpatient | 2,191 | 7.1% |

**Table 2. Distribution of extracted labels for PE cases.**

| Characteristic | Subgroup | Total (prevalence%) |
|---|---|---|
| **Laterality** | Bilateral | 1550 (47.3) |
| | Right | 1243 (37.9) |
| | Left | 447 (13.6) |
| | Unspecified | 39 (1.2) |
| **Acuity** | Acute | 2887 (88.1) |
| | Non_acute | 392 (11.9) |
| **Maximum Depth of Thrombosis** | Saddle | 147 (4.5) |
| | Main Lobar | 573 (17.5) |
| | Lobar | 871 (26.6) |
| | Segmental | 1315 (40.1) |
| | Subsegmental | 315 (9.6) |
| | Unspecified | 58 (1.8) |
| **Pulmonary Infarction** | Absent | 2507 (76.5) |
| | Present/Possible | 772 (23.5) |
| **Right Heart Strain** | Absent | 2114 (64.5) |
| | Present/Possible | 1165 (35.5) |
| **Pulmonary Hypertension** | Absent | 2662 (81.2) |
| | Present/Possible | 617 (18.8) |

**Table 3.** Aidoc Model Performance across demographic subgroups

| | Sensitivity (%) | Specificity (%) | PPV (%) | NPV (%) | F1 Score | Accuracy (%) | PE Cases (Prevalence) |
|---|---|---|---|---|---|---|---|
| **All Patients** | 86.80% (85.58% - 87.92%) | 99.14% (99.03% - 99.25%) | 92.32% (91.40% - 93.27%) | 98.43% (98.28% - 98.57%) | 0.8947 (0.8869 - 0.9028) | 97.82% (97.66% - 97.99%) | 3279 (10.69%) |
| **Patient Class** | | | | | | | |
| **Emergency** | 87.27% (85.78% - 88.76%) | 99.25% (99.12% - 99.38%) | 92.18% (90.94% - 93.43%) | 98.72% (98.55% - 98.87%) | 0.8965 (0.8865 - 0.9074) | 98.15% (97.97% - 98.33%) | 2031 (9.20%) |
| **Inpatient** | 86.45% (84.27% - 88.55%) | 98.80% (98.53% - 99.10%) | 93.42% (91.85% - 95.01%) | 97.37% (96.92% - 97.80%) | 0.8980 (0.8838 - 0.9122) | 96.77% (96.31% - 97.20%) | 1052 (16.43%) |
| **Outpatient** | 83.69% (78.48% - 88.57%) | 98.85% (98.34% - 99.30%) | 87.71% (82.94% - 92.46%) | 98.40% (97.86% - 98.91%) | 0.8562 (0.8182 - 0.8930) | 97.49% (96.85% - 98.13%) | 196 (8.95%) |
| **Race** | | | | | | | |
| **Asian** | 85.21% (74.60% - 93.75%) | 98.70% (97.94% - 99.34%) | 78.86% (68.57% - 88.24%) | 99.15% (98.56% - 99.71%) | 0.8179 (0.7377 - 0.8858) | 97.97% (97.06% - 98.75%) | 61 (5.43%) |
| **Black** | 85.93% (84.44% - 87.46%) | 99.18% (99.04% - 99.32%) | 92.90% (91.69% - 94.07%) | 98.26% (98.06% - 98.46%) | 0.8928 (0.8825 - 0.9033) | 97.71% (97.50% - 97.93%) | 1971 (11.11%) |
| **Other** | 83.20% (74.24% - 91.67%) | 99.21% (98.55% - 99.67%) | 89.35% (80.87% - 96.00%) | 98.66% (97.89% - 99.34%) | 0.8608 (0.7939 - 0.9185) | 98.02% (97.10% - 98.86%) | 71 (7.37%) |
| **Unknown** | 88.48% (81.60% - 94.74%) | 99.10% (98.42% - 99.74%) | 91.84% (85.42% - 97.33%) | 98.68% (97.81% - 99.48%) | 0.9008 (0.8518 - 0.9457) | 98.01% (96.97% - 98.95%) | 88 (10.27%) |
| **White** | 88.51% (86.53% - 90.40%) | 99.10% (98.90% - 99.29%) | 92.34% (90.70% - 93.97%) | 98.60% (98.35% - 98.86%) | 0.9038 (0.8907 - 0.9175) | 97.95% (97.69% - 98.24%) | 1088 (10.90%) |
| **Sex** | | | | | | | |
| **Female** | 87.08% (85.48% - 88.74%) | 99.24% (99.10% - 99.37%) | 92.66% (91.43% - 93.84%) | 98.59% (98.40% - 98.77%) | 0.8978 (0.8873 - 0.9086) | 98.04% (97.84% - 98.24%) | 1858 (9.89%) |
| **Male** | 86.42% (84.64% - 88.25%) | 98.96% (98.76% - 99.14%) | 91.86% (90.33% - 93.25%) | 98.17% (97.91% - 98.43%) | 0.8905 (0.8778 - 0.9023) | 97.46% (97.20% - 97.74%) | 1421 (11.95%) |
| **Age Group** | | | | | | | |
| **18-40** | 87.44% (84.21% - 90.62%) | 99.35% (99.11% - 99.54%) | 91.16% (88.09% - 93.81%) | 99.03% (98.77% - 99.29%) | 0.8925 (0.8698 - 0.9129) | 98.49% (98.17% - 98.80%) | 399 (7.15%) |
| **40-65** | 85.12% (83.14% - 87.05%) | 99.20% (99.04% - 99.36%) | 92.44% (91.02% - 93.84%) | 98.30% (98.06% - 98.54%) | 0.8863 (0.8725 - 0.8995) | 97.75% (97.48% - 98.01%) | 1302 (10.32%) |
| **65+** | 88.00% (86.38% - 89.68%) | 98.97% (98.77% - 99.16%) | 92.55% (91.16% - 93.85%) | 98.27% (98.03% - 98.52%) | 0.9021 (0.8903 - 0.9130) | 97.59% (97.32% - 97.84%) | 1578 (12.65%) |
| **Ethnicity** | | | | | | | |
| **Hispanic** | 83.56% (76.31% - 89.83%) | 99.30% (98.84% - 99.69%) | 91.07% (85.18% - 96.23%) | 98.61% (97.93% - 99.16%) | 0.8710 (0.8190 - 0.9156) | 98.07% (97.37% - 98.72%) | 110 (7.82%) |
| **Not-Hispanic** | 86.58% (85.30% - 87.69%) | 99.14% (99.02% - 99.25%) | 92.31% (91.33% - 93.29%) | 98.40% (98.24% - 98.55%) | 0.8935 (0.8852 - 0.9016) | 97.79% (97.61% - 97.97%) | 3002 (10.71%) |
| **Unknown** | 92.88% (88.82% - 96.90%) | 98.89% (98.22% - 99.45%) | 92.77% (88.82% - 96.32%) | 98.91% (98.26% - 99.54%) | 0.9281 (0.8986 - 0.9565) | 98.09% (97.29% - 98.80%) | 167 (13.33%) |

**Table 4.** Model performance across different PE subgroups.

| Category | Subgroup | Sensitivity % (95% CI) | PE Cases (Prevalence) |
|---|---|---|---|
| **Acuity** | Acute | **89.70 (88.64 – 90.75)** | 2887 (88.05%) |
| | Non_acute | 65.31 (60.71 – 69.90) | 392 (11.95%) |
| **Maximum Level of Thrombosis** | Saddle | **99.31 (97.96 – 100)** | 147 (4.47) |
| | Main Lobar | 91.15 (89.09 – 93.00) | 871 (26.56) |
| | Segmental | 82.32 (80.23 – 84.41) | 1315 (40.10) |
| | Subsegmental | 72.92 (67.93 – 77.46) | 315 (9.61) |
| **Laterality** | Bilateral | 95.24 (94.13 – 96.32) | 1550 (47.27%) |
| | Right | 82.21 (80.13 – 84.23) | 1243 (37.91%) |
| | Left | 70.54 (66.66 – 74.72) | 447 (13.63%) |
| **Pulmonary Infarction** | Absent | 85.48 (84.04 – 86.84) | 2507 (76.46) |
| | Present/Possible | 90.92 (88.86 – 92.88) | 772 (23.54) |
| **Right Heart Strain** | Absent | 82.16 (80.60 – 83.68) | 2114 (64.47) |
| | Present/Possible | 95.19 (93.82 – 96.48) | 1165 (35.53) |
| **Pulmonary Arterial Hypertension** | Absent | 86.76 (85.46 – 88.05) | 2662 (81.18) |
| | Present/Possible | 86.91 (84.28 – 89.63) | 617 (18.82) |

**Table 5.** Performance analysis based on intersection of “Age” and “Patient Class” subgroups.

| Age | Clinical setting | PE cases/Total (prevalence) | sensitivity | specificity |
|---|---|---|---|---|
| **<40** | Emergency | 265/4366 (6.1%) | 87.9% (83.8-91.7) | **99.4% (99.2-99.7)** |
| | Inpatient | 116/939 (12.4%) | 85.4% (77.9-91.7) | 99.0% (98.3-99.6) |
| | Outpatient | 18/276 (6.5%) | 94.4% (81.8-100) | 99.2% (98.0-100) |
| **40-65** | Emergency | 837/9273 (9.0%) | 86.3% (83.9-88.6) | 99.3% (99.1-99.5) |
| | Inpatient | 382/2382 (16.0%) | 84.2% (80.3-87.8) | 98.7% (98.2-99.2) |
| | Outpatient | 83/963 (8.6%) | 77.1% (67.4-86.2) | 99.2% (98.5-99.7) |
| **65+** | Emergency | 929/8444 (11.0%) | 88.0% (86.0-89.9) | 99.1% (98.9-99.3) |
| | Inpatient | 554/3082 (18.0%) | 88.3% (85.7-90.9) | 98.8% (98.4-99.2) |
| | Outpatient | 95/952 (10.0%) | 87.5% (80.5-93.9) | 98.4% (97.5-99.2) |